\documentclass[journal]{IEEEtran}
\usepackage{cite}
\usepackage{url}
\usepackage{amsmath,amssymb} 
\usepackage{algorithm}
\usepackage{algorithmic}
\usepackage{graphics}
\usepackage[normalem]{ulem}
\usepackage{multirow}
\usepackage{float}
\usepackage{rotating}
\usepackage[table]{xcolor}
\usepackage[pagebackref=false,breaklinks=true,colorlinks, bookmarks=false]{hyperref}

\usepackage{tikz}
\usepackage{pgfplots}
\pgfdeclarelayer{background}
\pgfdeclarelayer{foreground}
\pgfsetlayers{background,main,foreground}

\usepackage{marginnote}

\usepackage{silence}
\ifCLASSINFOpdf
\else
\fi
\def\ie{\emph{i.e.}}

\definecolor{bblue}{rgb}{0,150,230}
\definecolor{mygray}{gray}{.92}

\newcommand{\tr}[1]{\textbf{\textcolor{red}{#1}}}

\newcommand{\tb}[1]{\textcolor{blue}{#1}}

\graphicspath{{./Imgs/}{../../figure/}}

\begin{document}
%
\title{A Benchmark for Studying Diabetic Retinopathy: Segmentation, Grading, and Transferability}
%
%
%

\author{Yi Zhou, \textit{IEEE Member}, Boyang Wang, Lei Huang, Shanshan Cui, and Ling Shao, \textit{IEEE Senior Member}
\thanks{Corresponding author: \textit{Yi Zhou}.}
\thanks{Y.~Zhou, B.~Wang, L.~Huang, S.~Cui, and L.~Shao are with the Inception Institute of Artificial Intelligence, Abu Dhabi, UAE. (e-mails: yizhou.szcn@gmail.com)}
}

\markboth{Preprint to IEEE Transactions on Medical Imaging, 2020}%
{Zhou \MakeLowercase{\textit{et al.}}: FGADR}
%

\maketitle

\begin{abstract}

People with diabetes are at risk of developing an eye disease called diabetic retinopathy (DR). This disease occurs when high blood glucose levels cause damage to blood vessels in the retina. Computer-aided DR diagnosis has become a promising tool for the early detection and severity grading of DR, due to the great success of deep learning. However, most current DR diagnosis systems do not achieve satisfactory performance or interpretability for ophthalmologists, due to the lack of training data with consistent and fine-grained annotations. To address this problem, we construct a large fine-grained annotated DR dataset containing 2,842 images (FGADR). Specifically, this dataset has 1,842 images with pixel-level DR-related lesion annotations, and 1,000 images with image-level labels graded by six board-certified ophthalmologists with intra-rater consistency. The proposed dataset will enable extensive studies on DR diagnosis. Further, we establish three benchmark tasks for evaluation: 1. DR lesion segmentation; 2. DR grading by joint classification and segmentation; 3. Transfer learning for ocular multi-disease identification. Moreover, a novel inductive transfer learning method is introduced for the third task. Extensive experiments using different state-of-the-art methods are conducted on our FGADR dataset, which can serve as baselines for future research. Our dataset will be released in \url{https://csyizhou.github.io/FGADR/}.

\end{abstract}


\section{Introduction}\label{sec:introduction}

\IEEEPARstart{D}{iabetic} retinopathy (DR) is a type of ocular disease caused by high levels of blood glucose and high blood pressure, which can damage the blood vessels in the back of the eye (retina) and lead to blindness. One-third of people living with diabetes have some degree of diabetic retinopathy, and every person who has diabetes is at risk of developing it. Accurately grading diabetic retinopathy is time-consuming for ophthalmologists and can be a significant challenge for beginner ophthalmology residents. Therefore, developing an automated diagnosis system for diabetic retinopathy has significant potential benefits.

\begin{figure}[t]
\begin{center}
\includegraphics[width=1.0\linewidth]{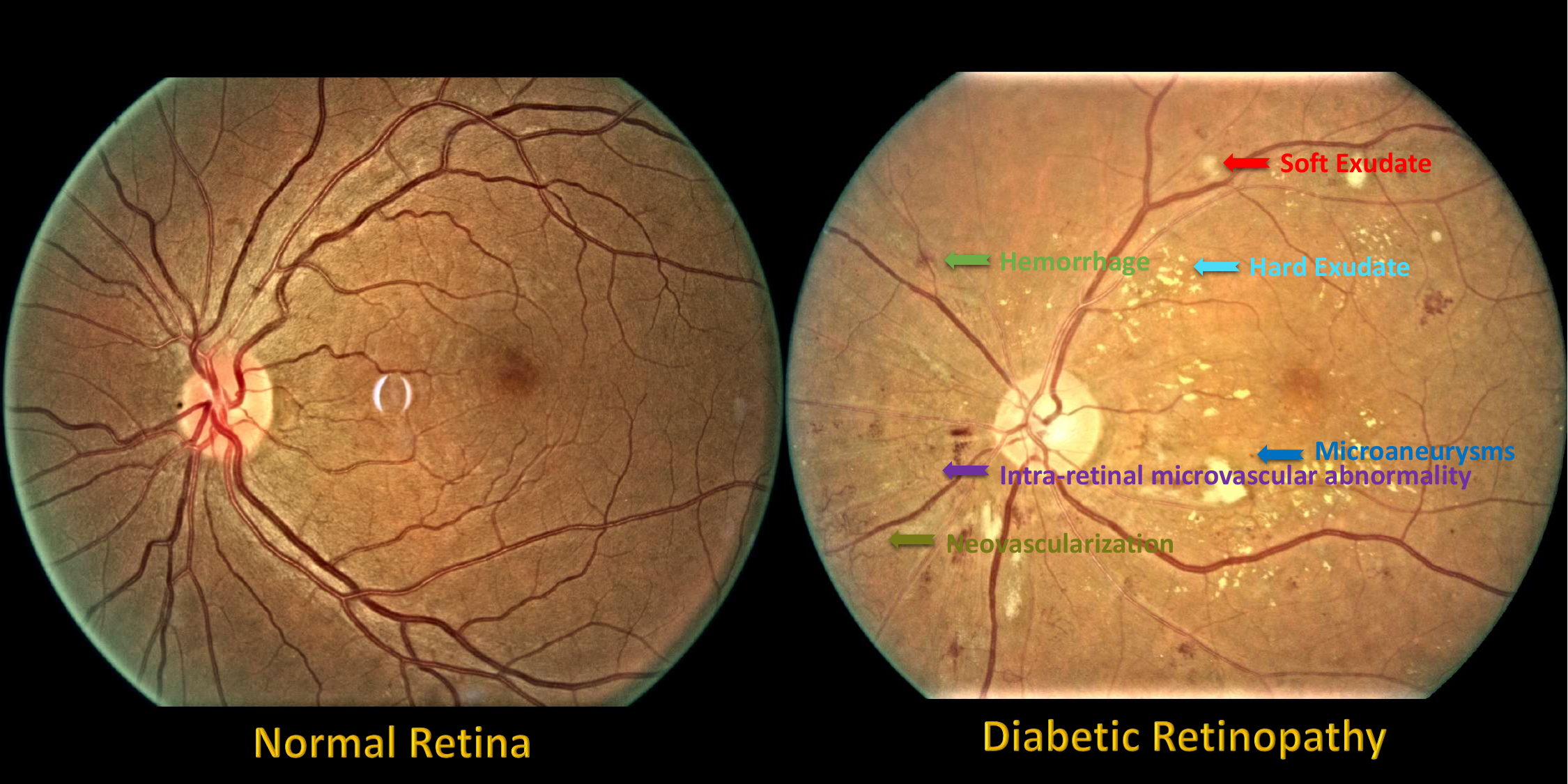}
\end{center}
   \caption{Illustration of diabetic retinopathy retina. The left image shows a normal retina, while the right one is a DR-4 retina.}
\label{fig:dr_intro}
 \vspace{-2ex}
\end{figure}

According to international protocol \cite{gulshan2016development,drgrading}, the severity of DR can be graded into five stages (0-4): no retinopathy (0), mild non-proliferative DR (NPDR) (1), moderate NPDR (2), severe NPDR (3), and proliferative DR (4). The grading usually depends on the number and size of different related lesion appearances and complications. Figure~\ref{fig:dr_intro} provides two examples, comparing a normal and a diabetic retinopathy retina containing multiple lesions. For example, microaneurysms (MAs) are the earliest clinically visible evidence of DR. These are local capillary dilatations that appear as small red dots. Moderate NPDR contains `dot' or `blot' shaped hemorrhages (HEs) in addition to microaneurysms. Hard exudates (EXs) are distinct yellow-white intra-retinal deposits which can vary from small specks to larger patches. They are principally observed in the macular region, as the lipids coalesce and extend into the fovea. Soft exudates (SE), also sometimes referred to as `cotton-wool spots' (CWS), are greyish-white patches of discoloration in the nerve fiber layer, or pre-capillary arterial occlusions. They usually appear in severe DR stages. Moreover, intra-retinal microvascular abnormalities (IRMAs) are areas of capillary dilatation and new intra-retinal vessel formation. A pre-proliferate DR stage can be predicted once IRMA is present in numbers. Neovascularization (NV) is a significant factor of proliferate DR. As the retina becomes more ischaemic, new blood vessels may arise from the optic disc or in the periphery of the retina. Therefore, identifying these related regions can be helpful for DR grading.

Over the past decade, computer vision and deep learning based algorithms have been largely explored to contribute to the medical imaging research community. With successful developments in deep convolutional neural networks (CNNs), image classification \cite{irvin2019chexpert}, object detection \cite{yan2018deeplesion}, semantic segmentation \cite{fan2020inf}, and image synthesis \cite{shin2018medical} frameworks, have all been investigated to analyze medical images for addressing different tasks. To study diabetic retinopathy \cite{asiri2019deep}, most previous works can be coarsely categorized into three important branches. First, the most valuable task is to predict diabetic retinopathy progression (\ie~grading \cite{gulshan2016development,arcadu2019deep,gargeya2017automated,seoud2015automatic,jiang2019interpretable,tu2020sunet}). Gulshan \emph{et al.} \cite{gulshan2016development} adopted the Inception-v3 architecture to train a DR grading model, which aims to directly learn the local features rather than explicitly detecting lesions. In \cite{jiang2019interpretable}, an automated image-level DR grading system was built on an ensemble of multiple well-trained deep learning models. Some of these deep models were also combined with the AdaBoost to reduce the bias of each individual model. Second, lesion-based diabetic retinopathy detection \cite{yang2017lesion,wang2017zoom,lin2018framework,Zhou_2019_CVPR,seoud2016red,he2019dme,antal2012ensemble,tang2012splat,zhang2009hierarchical,zhou2020encoding} has also been investigated. Yang. \emph{et al.} \cite{yang2017lesion} proposed to integrate lesion detection and grading by designing two-stage deep convolutional neural networks. Specifically, a local network is first trained to classify the patches into different lesions, and then the second network predicts the severity grades of DR. In \cite{wang2017zoom}, a zoom-in-net was proposed to learn attention maps which highlight abnormal regions, and then provides the grading levels of DR in both global and local manners. Third, several image generation methods \cite{zhou2019high,niu2018pathological,zhao2018synthesizing,costa2017towards} have been proposed for synthesizing retinal images. This technique can be used for data augmentation to address imbalances in DR training data. Niu \emph{et al.} \cite{niu2018pathological} proposed to synthesize fundus images given the pathological descriptors and vessel segmentation masks. DR-GAN, proposed in \cite{zhou2019high}, attempts to generate high-resolution retinal images with different grade levels by manipulating arbitrary grading and lesion information.

Currently, the two biggest obstacles to the progress of computer-aided diagnosis systems for DR are limited amounts of training data and inconsistent annotations. While there are a few public DR databases, such as \cite{kaggle-eyepacs,decenciere2014feedback,porwal2018indian,staal2004ridge}, but most of them only contain image-level labels, and annotations are often inaccurate. Constructing a large dataset with high-quality and fine-grained annotations would significantly contribute to research in DR diagnosis. For example, pixel-level annotations of DR-related lesions are highly beneficial for developing lesion-based segmentation models, as well as for training more interpretable grading models for ophthalmologists. Moreover, if fine-grained annotations of numerous lesions are provided, this rich information can be used to improve the ability of representation learning, as well as enable the models to be transferred for other ocular disease identification tasks without annotations. Therefore, in this paper, we propose a new benchmark for studying diabetic retinopathy diagnosis systems. A large pixel-level annotated DR dataset is introduced, and three tasks are set up to evaluate different methods. \textbf{The main contributions of this benchmark work are as follows:}

\textbf{1. } We construct a DR dataset with fine-grained annotations, named FGADR, containing 1,842 fundus images with both pixel-level lesion annotations and image-level grading labels, and 1,000 images with only grading labels. Based on this dataset, algorithms such as semantic segmentation, image classification, transfer learning, supervised, and semi-supervised learning, can be extensively explored to advance research in the DR, and even more general, medical imaging community.

\textbf{2. } Three tasks are established to evaluate different methods on our newly proposed dataset. Extensive experiments and analyses are conducted. First, medical image segmentation methods are explored based on the pixel-level lesion annotations. Second, joint classification and segmentation frameworks are studied to improve the DR grading performance by exploiting more interpretable lesion segmentation results. Moreover, transfer learning for other ocular multi-disease identification is also investigated using our dataset.

\textbf{3. } To evaluate the third task, we also propose a novel inductive transfer learning method to improve the performance of ocular multi-disease identification. Multi-scale transfer connections and a domain-specific adversarial adaptation module are designed to bridge the task learning between the source and target domains. Experiments are conducted on our FGADR dataset and the ODIR-5K dataset \cite{odir2019}.

\begin{figure*}[t]
\begin{center}
\includegraphics[width=1.0\linewidth]{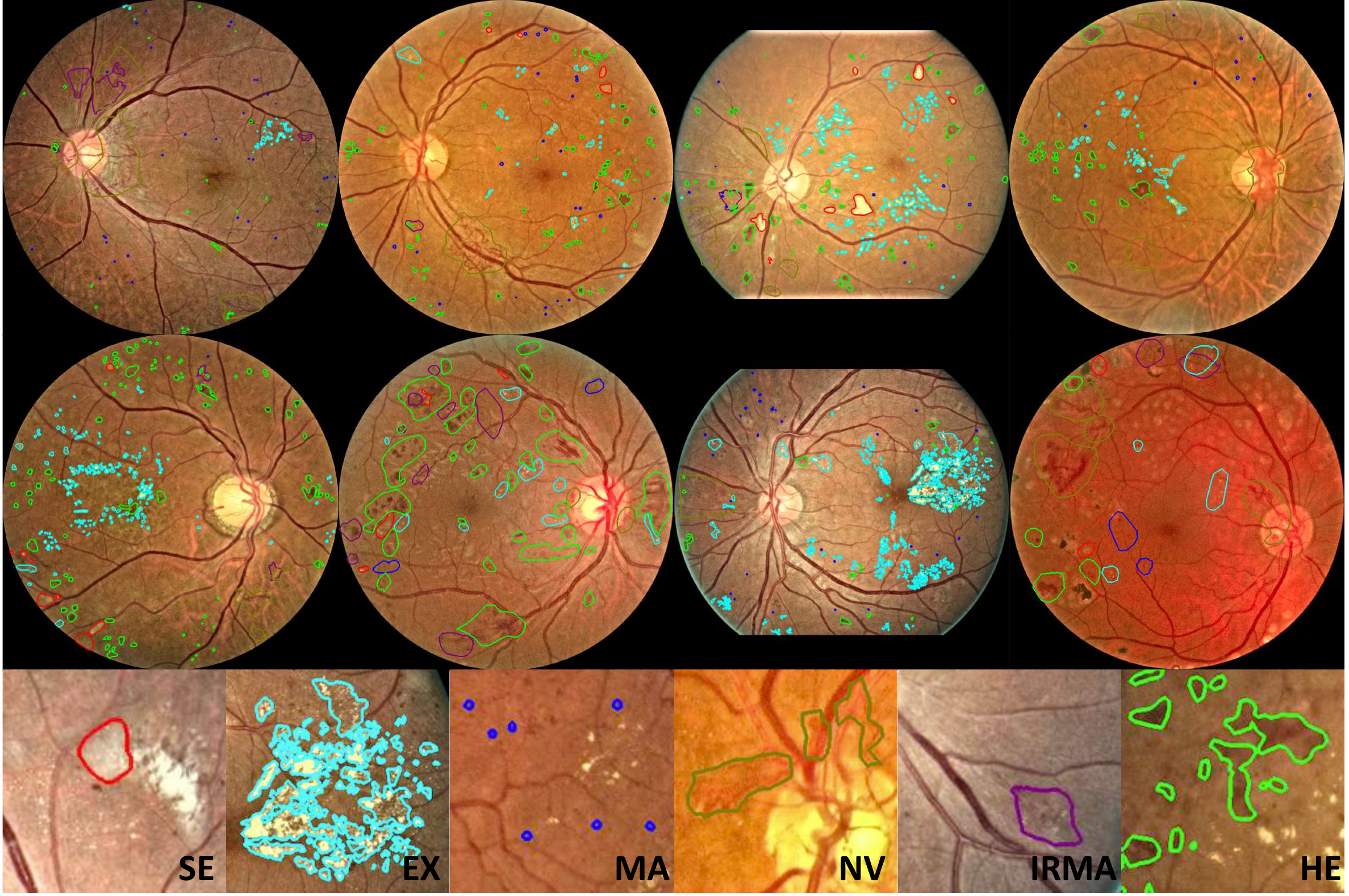}
\end{center}
   \caption{Pixel-level annotation examples from our FGADR dataset, including six different lesions. The blue, green, red, cyan, purple, and olive denote microaneurysms, hemorrhages, soft exudates, hard exudates, intra-retinal microvascular abnormalities, and neovascularization, respectively.}
\label{fig:demonstration}
\end{figure*}

\section{Datasets}\label{sec:datasets}

Most of the existing DR datasets only have image-level grading labels, with providing few pixel-level lesion-based annotations. A summary of some commonly used datasets related to DR is provided in Table~\ref{tab:DatasetSummary}. Models trained on these datasets can only be used to predict a severity grade without providing any interpretability for ophthalmologists as to why a fundus image is graded as a certain level. Therefore, one of the main goals of our benchmark is to introduce a large fine-grained annotated dataset for more explainable diagnosis of DR. Detailed information of existing datasets and our proposed dataset are as follows.

\begin{table}[t]
  \centering
  \footnotesize
  \renewcommand{\arraystretch}{1.5}
  \setlength\tabcolsep{0.5pt}
    \caption{A summary of public Diabetic Retinopathy imaging datasets.
    }\label{tab:DatasetSummary}
  \begin{tabular}{l|c|c|c}
    \rowcolor{mygray}
  	\hline
  	Dataset                                        &      Annotation modes       &         Images          &     Tasks     \\ \hline\hline
Kaggle - EyePACS \cite{kaggle-eyepacs}     &     Image-level     &      88,702     &     DR grading 0-4     \\
Kaggle - APTOS2019 \cite{kaggle-aptos2019}     &     Image-level      &     5,590      &    DR grading 0-4      \\
ODIR-5K \cite{odir2019}     &     Image-level      &      7,000     &     Multi-disease classification     \\
Messidor \cite{decenciere2014feedback}     &    Image-level       &      1,200     &     DR grading 0-3     \\ \hline
DRIVE \cite{staal2004ridge}     &     Pixel-level     &     40      &    Vessel segmentation      \\
IDRiD \cite{porwal2018indian}     &    Pixel-level      &    81       &    Segmentation \& Grading     \\
FGADR      &     Pixel-level     &      2,842     &    Segmentation \& Grading       \\ \hline
  \end{tabular}
\end{table}

\subsection{Existing DR Grading Datasets}

\subsubsection{Kaggle-EyePACS \cite{kaggle-eyepacs}} This consists of 35,126 training images and 53,576 testing images only containing grading labels. The images are collected from different sources with various lighting conditions and weak annotation quality. The presence of DR in each image is rated on a scale of 0 to 4. In this dataset, some images contain artifacts, are out of focus, underexposed, or overexposed.

\subsubsection{Kaggle-APTOS2019 \cite{kaggle-aptos2019}} This consists of 3,662 training images and 1,928 testing images, also with grading labels only. This dataset also suffers from noise in both images and labels.

\subsubsection{ODIR-5K \cite{odir2019}} This is a structured ophthalmic dataset of 5,000 patients. Multi-label image-level annotations for eight eye disease categories, including diabetes, glaucoma, cataract, age-related macular degeneration (AMD), hypertension, myopia, normal, and other diseases, are provided. Each patient may contain one or more disease labels. We adopt this dataset in the last task to explore transfer learning from DR to ocular multi-disease identification.

\subsubsection{Messidor \cite{decenciere2014feedback}} This contains 1,200 eye fundus images but its DR grading scale is different from those of previous datasets, having only four levels (0 to 3). In addition to DR grading, the risk of macular edema is also provided for each image with grading labels 0 to 2.

\subsection{Existing DR Lesion Segmentation Datasets}

\subsubsection{IDRiD \cite{porwal2018indian}} This dataset provides expert annotations of typical diabetic retinopathy lesions and normal retinal structures. The full set contains 516 images, but only 81 of them are labeled with pixel-level binary lesion masks. Abnormalities associated with DR, such as microaneurysms, hemorrhages, soft exudates and hard exudates, are provided.

\subsubsection{DRIVE \cite{staal2004ridge}} This dataset is used for evaluating the segmentation of blood vessels in retinal images, and contains pixel-level binary vessel masks. The 40 images are divided into a training and a testing set, each containing 20 images.

\begin{figure}[]
\begin{center}
\includegraphics[width=1.0\linewidth]{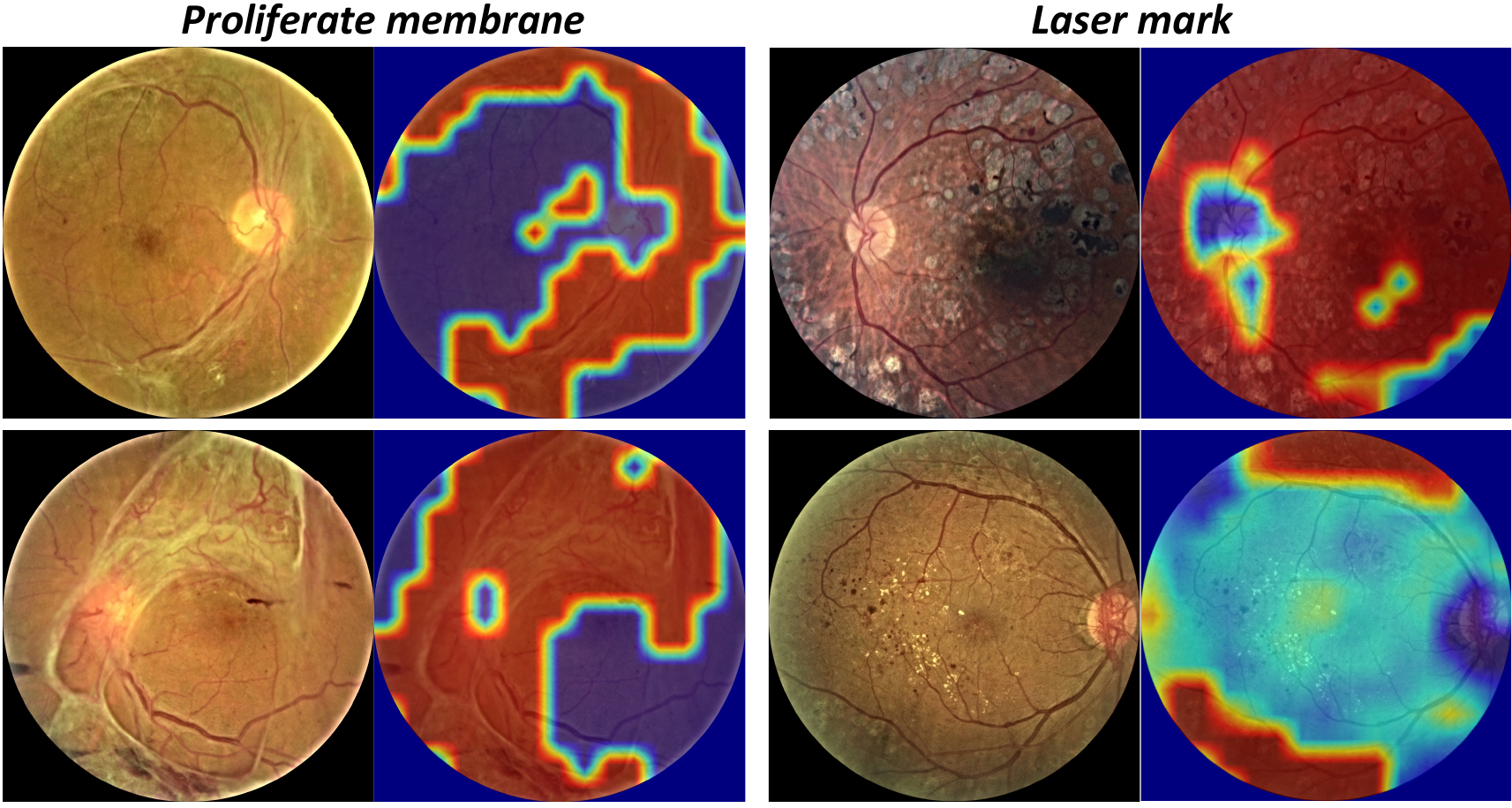}
\end{center}
   \caption{Examples of laser marks and proliferate membranes, and corresponding class activation maps by \cite{zhou2016learning}.}
\label{fig:las_mem}
\end{figure}

\begin{figure*}[t]
\begin{center}
\includegraphics[width=1.0\linewidth]{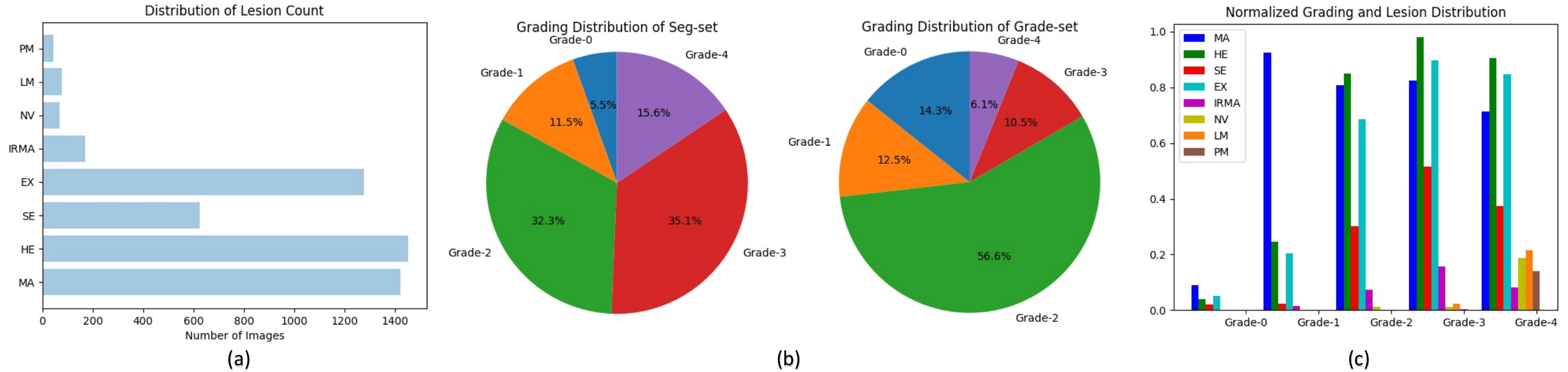}
\end{center}
   \caption{Statistics of our FGADR dataset. (a) Number of images for each pixel-level annotated lesion. (b) The left and right pie charts illustrate the grading distribution of the Seg-set and Grade-set, respectively. (c) Lesion distribution normalized by the number of images for different grades.}
\label{fig:statistics}
\end{figure*}

\subsection{Our FGADR Dataset}

We collected a fine-grained annotated diabetic retinopathy (FGADR) dataset, which consists of two sets. The first set, named Seg-set, contains 1,842 images with both pixel-level lesion annotations and image-level grading labels. The lesions include microaneurysms (MA), hemorrhages (HE), hard exudates (EX), soft exudates (SE), intra-retinal microvascular abnormalities (IRMA), and neovascularization (NV). The grading labels are annotated by three ophthalmologists. The second set, named Grade-set, is a set of 1,000 images with grading labels annotated by six ophthalmologists. This set is specifically designed for evaluating grading performance due to its high annotation confidence.

In addition to the six pixel-level lesions annotated in Seg-set, we also annotate the laser mark (LM) and proliferate membrane (PM) lesions. Laser marks and proliferate membranes are important lesions that usually appear in severe DR grades (i.e. grade-3 and grade-4). However, they appear as are global-like features, making them difficult to annotate in a pixel-wise manner. Thus, only image-level labels for these two lesions are provided, which indicate whether or not an image has the lesion. Some examples of these two lesions, as well as their class activation maps extracted by the weakly-supervised method \cite{zhou2016learning}, are shown in Figure~\ref{fig:las_mem}.

\subsubsection{Dataset Construction and Labeling}
The fundus image data were mainly collected from our local partner hospitals. To protect patient privacy, personal information was anonymized in our dataset construction. During data pre-cleaning, we only selected the best quality image for each patient ID. Thus, no two images in the dataset have the same retinal structure in terms of vessel or optic disk. This filtering ensures lesion diversity in FGADR. Moreover, since our main goal was to build a dataset for annotating pixel-level DR lesions, we preferred to select fundus images of high DR severity levels containing more lesions. Thus, we trained a DR grading model based on the Kaggle-EyePACS dataset \cite{kaggle-eyepacs}, and then applied it to our data collected from hospitals. We selected a set of high-quality images graded with DR levels of 2, 3, and 4 by the model, which might also contain misclassified grade-0 and grade-1 images inside, for annotation. Three ophthalmologists (two resident physicians and one physician-in-charge) were invited to annotate this Seg-set. The resident ophthalmologists carried out the preliminary annotation, and the physician-in-charge took responsibility for the final verification. Some annotation examples are provided in Figure~\ref{fig:demonstration}. In addition to the lesion annotation, image-level grading annotation for the Seg-set was also done, in a voting manner by the three ophthalmologists.

An extra set, the Grade-set, is also provided with grading labels only. The role of this set is to evaluate the performance of DR grading models. To ensure the accuracy of the grading annotations, we invited six ophthalmologists (three resident physicians, and three physicians-in-charge) for annotation, and again used a voting manner for the final labels.

\subsubsection{Annotation Criteria}
We employed a strict annotation criteria and the whole annotation process of the Seg-set of FGADR took over 10 months. We asked three ophthalmologists to strictly guarantee the annotation accuracy through a quality control process. Details: MAs appear as small red dots in the color photographs with staining on angiogram. If there is no angiogram, a red dot on the color photograph is graded as an MA if the grader believed the lesion is a MA.  Red dot-like lesions are usually graded as retinal HEs, not MAs. EXs are small white or yellowish white deposits with sharp margins. Often, they appear waxy, shiny, or glistening. MAs that appear as white dots with no blood vessels visible in the lumen are considered EXs. Superficial white, pale yellow-white or grayish-white areas with feathery edges, frequently showing striations parallel to the nerve fiber layer are SEs. NVs of the disc are characterized by the development of variable caliber vessels anterior to the optic nerve or retina. IRMAs are slightly larger in caliber with a broader arrangement and are always found within intraretinal layers. Moreover, the DR grading criteria strictly follows the international protocol \cite{drgrading}.

\subsubsection{Dataset Statistics}

\textbf{(a)} Most images in the Seg-set contain one or more kinds of lesions annotated. The distribution of lesion counts is shown in Figure~\ref{fig:statistics} (a). We observe that microaneurysms, hemorrhages, and hard exudates are the three most common lesions in DR images, while intra-retinal microvascular abnormalities, neovascularization, laser marks, and proliferate membranes rarely appear.

\textbf{(b)} The grading distributions of the Seg-set and Grade-set are illustrated in Figure~\ref{fig:statistics} (b). Since all the samples in the Seg-set are coarsely selected through a pre-trained grading model, the ratios of grade 0 and 1 are low. More specifically, Seg-set has 1,842 images ([`grade': the number of images] `0': 101, `1': 212, `2': 595, `3': 647, `4': 287), and Grade-set has 1000 images (`0': 143, `1': 125, `2': 566, `3': 105, `4': 61).

\textbf{(c)} We also illustrate various lesion distributions related to the five grading levels in Figure~\ref{fig:statistics} (c), with normalization. As shown, microaneurysms are the first DR lesions to appear usually starting in the early stages (grade-0 or grade-1). Moreover, the number of all lesions generally grows as the DR grading level increases. Although it is difficult to differentiate stages 3 and 4, only based on lesion distributions, we observe that neovascularization, laser marks, and proliferate membranes are good factors for further discrimination.

\section{Benchmark settings for DR lesion segmentation, grading, and transfer learning}

With the proposed FGADR dataset, we can explore various problems related to diabetic retinopathy, such as pixel-level lesion segmentation and image-level DR severity grading. We set up three tasks to evaluate different methods on our dataset. In Task 1, classic segmentation models for medical imaging are applied to multiple DR lesions. In Task 2, we investigate DR grading by joint classification and lesion segmentation, which we believe is a challenging and interesting research topic. Moreover, due to our large number of fine-grained annotations on fundus images, a transfer learning method is also proposed, in Task 3, to explore whether or not our dataset can contribute to the diagnosis of other eye diseases.

\subsection{Task 1: DR Lesion Segmentation}
Task 1 is designed to evaluate DR lesion segmentation models, where numerous pixel-level annotations are provided. This task is based on the Seg-set of our FGADR only. It contains six sub-tasks, including the segmentation of microaneurysms, soft exudates, hard exudates, hemorrhages, intra-retinal microvascular abnormalities, and neovascularization binary masks. For each sub-task, we conduct two-fold cross validation experiments, using 50\% of images for training and 50\% for testing.

\subsection{Task 2: Grading by Joint Classification and Segmentation}
Since one of the main goals of DR diagnosis is to rate the severity level from 0 to 4, we would also like to evaluate the performance of grading models on our Grade-set containing 1,000 test images. The grading task is implemented as a normal classification problem. We aim to combine the classification task with lesion segmentation to jointly contribute to the final diagnosis of DR. Image-level grading labels from Kaggle-EyePACS \cite{kaggle-eyepacs} and the Seg-set of our FGADR dataset are combined to train the classification models, while the pixel-level labels of the Seg-set are used for training the segmentation models. The overall framework of this task is to exploit the Seg-set data to train DR-related lesion segmentation modules and extract DR-related lesion features on data of Kaggle-EyePACS and the Grade-set of FGADR for joint learning and evaluation of the grading models. To learn grading models, the features extracted by the segmentation branch (trained using pixel-level DR-related lesion annotations) are integrated with those obtained by the grading branch (trained using only image-level DR grading labels) to improve the results.

Several works on joint classification and segmentation models have been proposed. For instance, \cite{lin2018framework} introduced a lesion detection model to first extract lesion information, and then used an attention-based network to fuse original images and lesion features to identify DR. A collaborative learning framework was introduced in \cite{Zhou_2019_CVPR} to optimize a lesion segmentation model and a disease grading model in an end-to-end fashion. Then, a lesion attentive classification module was proposed to improve the severity grading accuracy, and a lesion attention module to refine lesion maps extracted from unannotated data for semi-supervised segmentation. Moreover, in \cite{wu2020jcs}, segmentation and classification are conducted in parallel. The predicted lesion probability maps from the segmentation model, and the class activation maps from the weakly-supervised classification model, are combined for joint diagnosis. In this task, we adopt the above three methods as baselines to evaluate the DR grading performance, and explore how the grading model can benefit from learning the lesion segmentation model trained on our data. Moreover, the image-level laser mark and proliferate membrane lesion labels are also additionally used to co-train the classification models.

\subsection{Task 3: Inductive Transfer Learning for Ocular Multi-Disease Identification}

In addition to diagnosing diabetic retinopathy, we also want to explore whether our fine-grained annotated dataset can benefit learning other eye disease identification tasks. First, some eye diseases have similar lesion appearances to DR. For example, AMD is an acquired degeneration of the retina that has abnormalities such as neovascular derangement and hemorrhages. Hypertensive retinopathy usually contains exudates and hemorrhages. These shared lesions can be used to help train the corresponding disease identification models without pixel-level annotations. Second, the rich annotations in our dataset can also enhance the generalization ability of models in terms of representation learning on fundus images, since various textures and colors are well delineated. Therefore, we propose a transfer learning method to improve multi-disease identification performance using our dataset. The evaluation is conducted on the ODIR-5K \cite{odir2019} dataset.

Transfer learning involves using knowledge learned from tasks for which a lot of labeled data is available in settings with limited labeled data. It can be coarsely categorized into three branches, based on different situations. First, regardless of whether the source and target domains are similar or not, if the tasks are different, inductive transfer learning \cite{pan2009survey} is used. In contrast, if the source and target domains are different but the task is the same, transductive transfer learning \cite{arnold2007comparative} is preferred. Moreover, if both the domains and tasks are different, unsupervised transfer learning \cite{chang2017unsupervised} needs to be considered. In our case, an inductive transfer learning method is required, since both the source and target domains are fundus images, but the source and target domain tasks are DR lesion segmentation and multi-disease classification, respectively. The inductive transfer learning algorithms try to utilize the inductive biases of the source domain to help improve the target task. Depending upon whether the source domain contains labeled data or not, this strategy can be further divided into two subcategories, similar to multitask learning and self-taught learning, respectively.

\begin{figure}[]
\begin{center}
\includegraphics[width=1.0\linewidth]{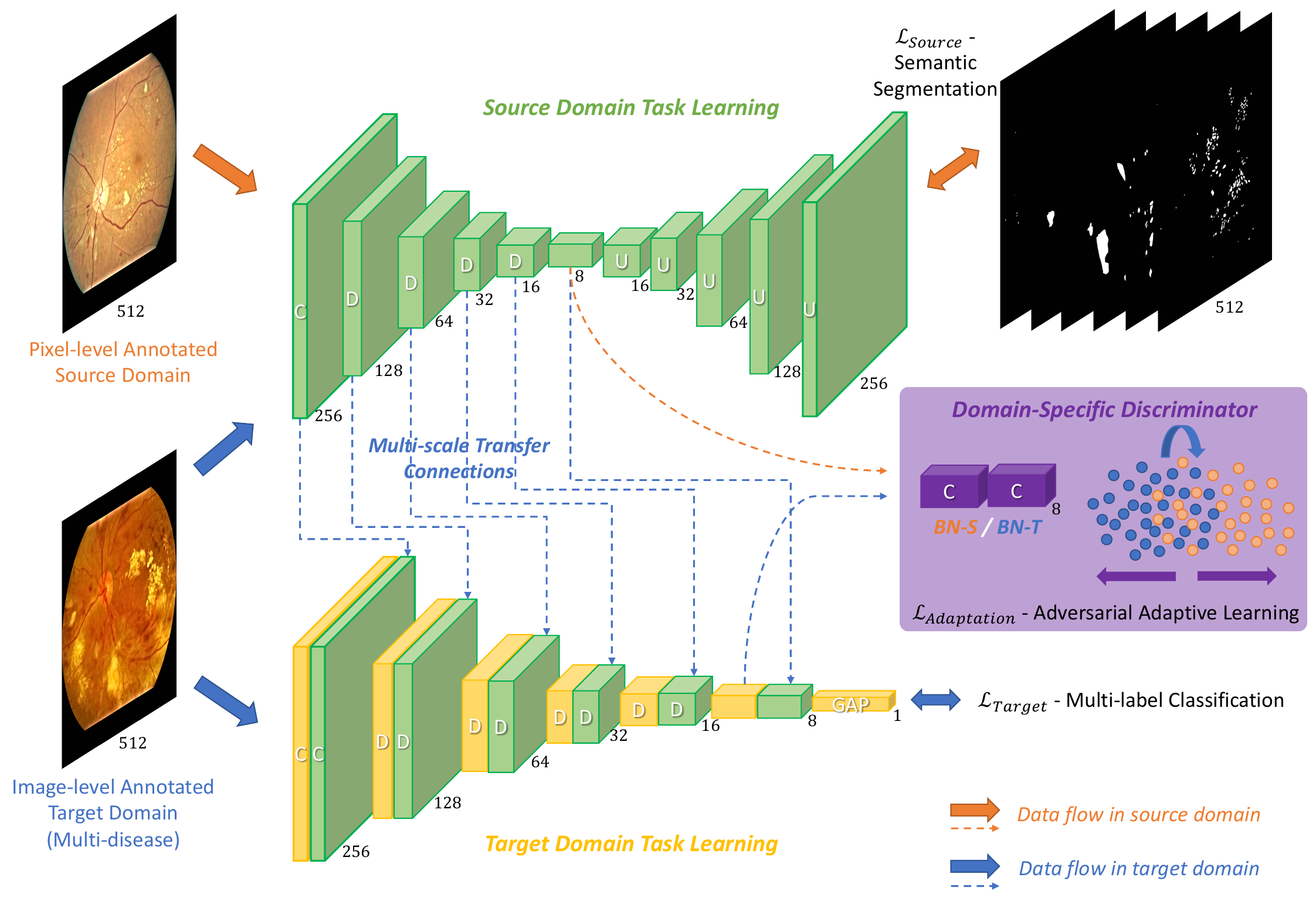}
\end{center}
   \caption{Overview of our proposed inductive transfer learning method for multi-disease identification. $C$ denotes a convolutional layer. $D$ denotes a dense block. A transition layer is adopted after each $D$. $U$ consists of an upsampling operation and a convolutional layer. GAP is global average pooling. BN-S and BN-T denote the separate batch normalization of the source and target domain.}
\label{fig:transfer}
\end{figure}

Our proposed inductive transfer learning method, which consists of three modules, is illustrated in Figure~\ref{fig:transfer}. First, the source domain task is to learn a lesion segmentation module related to DR. Second, the target domain task is to learn a multi-label classification module for identifying various eye diseases. Basically, a multi-scale transfer connection (MTC) is proposed to extend the strong feature extraction ability learned from the source domain data to the target domain data. Thus, the combined feature representations for the target domain data are enhanced, particularly for encoding lesion appearances included in our FGADR dataset. Moreover, a domain-specific adversarial adaptation (DSAA) module is proposed to adapt the representation distribution of the target and source domain data, while maintaining the disease discrepancy, through the addition of a domain-specific discriminator. We introduce the DSAA because we aim to adapt the representations of the two domains so that the segmentation module trained on the source domain data can better fit the target domain data and extract more effective multi-scale transferred features. In other words, the DSAA is proposed to enhance the effectiveness of the MTC.

\textbf{Details of the proposed algorithm:}

Let $\mathcal{D}_{S}$ denote the source domain data and $Y_S$ denote the corresponding labels. $\mathcal{L}_{S}$ is the loss for learning the source domain task. Moreover, let $\mathcal{L}_{T}$ denote the target domain data and $Y_T$ denote the corresponding labels. $\mathcal{L}_{T}$ is the loss for learning the target domain task. Then, an additional adaptation loss $\mathcal{L}_{A}$ is also proposed to adapt the two domain distributions in an adversarial learning manner. We generalize the overall loss function as:
\begin{equation}\label{equ:loss}
  \mathcal{L} = \mathcal{L}_{S}(\mathcal{D}_{S}, Y_S) + \lambda \mathcal{L}_{T}(\mathcal{D}_{T}, Y_T) + \gamma \mathcal{L}_{A}(\mathcal{D}_{S}, \mathcal{D}_T),
\end{equation}
where $\lambda$ and $\gamma$ balance the weights of different loss parts.

For the lesion segmentation module, we simply adopt the Dense U-Net structure, introduced in \cite{li2018h}, as the source domain backbone without too many bells and whistles. Details are shown in Figure~\ref{fig:transfer}. A transition layer \cite{li2018h} is adopted after each dense block. Since our input size is two times bigger than that of \cite{li2018h}, we add one more transition layer after the last dense block in the encoder to suitably increase the reception field. To optimize this segmentation module for source domain data, a pair of input images and corresponding lesion masks are provided. $\mathcal{L}_{S}$ adopts the weighted binary cross-entropy loss and Dice loss, as in Task 1.

In the target domain, a similar DenseNet backbone is adopted for learning a multi-label classification module. We propose multi-scale transfer connections to integrate features learned from the segmentation module. As illustrated in Figure~\ref{fig:transfer}, given a target domain image, its multi-scale features are extracted by the encoder of the segmentation module. Then, these features are concatenated with the corresponding scale features in the classification module. Thus, the descriptive representations learned from the segmentation module can be transferred to the classification module only supervised with image-level labels. Moreover, $\mathcal{L}_{T}$ adopts the weighted binary cross-entropy loss.

Since there exists feature distribution difference between the source and target domain (introduced by the different data sources), we aim to adapt the representations of the two-domain data so that the segmentation module trained on source domain data can fit the target domain data and extract better multi-scale transferred features. Such transferred knowledge of disease patterns shared between the two domains can improve the results of the target domain task. Moreover, due to the disease discrepancy introduced by the target domain, domain-specific properties are considered in our method as well. Therefore, a DSAA method is proposed to address the domain adaptation. First, we extract the bottleneck feature vector from the segmentation module in the source domain, and the same-sized feature vector from the classification module in the target domain. Then, a domain-specific discriminator is proposed, which stacks two convolutional ($Conv$) layers to discriminate whether the features are from the source or target domain. 

In some previous works, domain-specific batch normalization (DSBN) \cite{chang2019domain} is adopted in the main network because all the convolutional layer parameters of the main network are shared between the source and target domains to learn domain-invariant features. This can be done because there is only a distribution shift in domain data structure introduced by the use of different data sources, while the tasks of the two domains are the same. However, in our task, we face not only a data distribution shift, but also disease discrepancy between the two domains.  As such, we do not share the main network parameters, but adopt separate branches to learn different tasks for the two domains. Thus, in this case, it is not appropriate to use DSBN in the main network for addressing the domain shift. Instead, we adopt DSBN to replace the standard batch normalization ($BN$) after each $Conv$ layer in the discriminator. The discriminator separates the branches of the $BN$ layers, using one for each domain, while sharing all the other $Conv$ parameters across domains. We adopt DSBN because we expect the domain-specific disease information within the discriminator to be removed effectively and increase the difficulty of training the discriminator, by exploiting the captured statistics and learned parameters from the given domain during adversarial learning $\mathcal{L}_{A}$ \cite{goodfellow2014generative}. Thus, the adversarial adaptation can constrain the encoders of the two domains to learn domain-invariant features, while maintaining the disease discrepancy. The domain-specific adaptation module is optimized with the two task learning modules, simultaneously.

\subsection{Evaluation Metrics}

To evaluate the segmentation performance in Task 1, we use four widely adopted metrics, \ie, the Dice Similarity Coefficient, Area Under the Curve of Receiver Operating Characteristic (AUC-ROC), Area Under the Curve of Precision-Recall (AUC-PR), and Mean Absolute Error (MAE). In our evaluation, we choose the $sigmoid$ function as the final prediction $S_p$. Thus, we measure the similarity/dissimilarity between the final the prediction map and pixel-level segmentation ground-truth $G$, which can be defined as follows:

\subsubsection{Dice Similarity Coefficient (Dice)} This is a classic metric for evaluating medical image segmentation. It is a region-based measure to evaluate the region overlap. We formulate it as:
\begin{equation}\label{equ:d_m}
  Dice = \frac{2 |S_p \cap G|}{|S_p| + |G|},
\end{equation}
\subsubsection{AUC-ROC} It compares the Sensitivity vs (1 - Specificity), in other words, compares the true positive rate versus false positive rate. The bigger the AUC-ROC, the greater the distinction between true positives and true negatives.

\subsubsection{AUC-PR} Precision-recall curves plot the positive predictive value against the true positive rate. Both the precision and recall focus on the positive class (the minority class) and are unconcerned with the true negatives (the majority class). Thus, when the data is imbalanced, PR is more suitable than ROC.

\subsubsection{Mean Absolute Error~(MAE)} This measures the pixel-wise error between $S_p$ and $G$, which is defined as:
\begin{equation}\label{equ:mae}
  \textit{MAE} = \frac{1}{w \times h} \sum_{x}^{w} \sum_{y}^{h} |S_p(x,y) - G(x,y)|.
\end{equation}

For Task 2, the DR grading performance is evaluated, as a five-grade classification problem. In addition to the classification confusion matrix and accuracy, the quadratic weighted kappa metric is adopted.

\subsubsection{Quadratic Weighted Kappa (Q.W.Kappa)} The quadratic kappa metric is the same as Cohen's kappa metric \cite{mchugh2012interrater} when weights are set to `Quadratic'. It is calculated as follows. First, a multi-class confusion $O$ is created between predicted and ground-truth ratings, followed by a weight matrix $w$ which calculates the weight between the ground-truth and predicted ratings. Then, the value counts for each rating in predictions and ground truths are calculated, and the outer product of two value count vectors is computed as $E$. Finally, $E$ and $O$ are normalized and used to calculate the weighted kappa.

To evaluate multi-disease classification performance in Task 3, Cohen's kappa, F-1 score, and AUC-ROC are used.

\subsubsection{Cohen's Kappa} This was proposed for agreement between two raters. The formulation is as follows:
\begin{equation}\label{equ:kappa}
  kappa = \frac{p_o - p_e}{1 - p_e},
\end{equation}
where $p_o$ and $p_e$ denote the relative observed agreement among raters and the hypothetical probability of chance agreement.

\subsubsection{F-1 Score} This is computed based on precision and recall rate, given by the following formula:
\begin{equation}\label{equ:f1}
  \textit{F1} = 2 \times \frac{Precision \times Recall}{Precision + Recall}.
\end{equation}
F-1 score keeps a balance between precision and recall. We use this comparison indicator if there is uneven class distribution, as precision and recall may give misleading results.

\begin{table*}[thp!]
\centering
  \footnotesize
  \renewcommand{\arraystretch}{1.35}
  \setlength\tabcolsep{1.0pt}
  \caption{\small Quantitative results of deep learning-based lesion segmentation models on our FGADR dataset. The two best results are shown in \tr{red} and \tb{blue}.
  }\label{tab:segmentation}
  \resizebox{1.0\textwidth}{!}{
\begin{tabular}{c|cccc|cccc|cccc|cccc|cccc|cccc}
\hline
\multirow{2}{*}{Methods} & \multicolumn{4}{c|}{MA} & \multicolumn{4}{c|}{HE} & \multicolumn{4}{c|}{EX} & \multicolumn{4}{c|}{SE} & \multicolumn{4}{c|}{IRMA} & \multicolumn{4}{c}{NV} \\ \cline{2-25} 
                         & Dice   &  ROC  &  PR  & MAE  & Dice  &  ROC  &   PR  & MAE &    Dice   &   ROC   &   PR   & MAE  & Dice   & ROC   &   PR  & MAE  & Dice   &   ROC  &   PR  & MAE &  Dice   &  ROC   &   PR  & MAE \\ \hline
FCN-8s \cite{long2015fully}                   &    0.468    &     0.925      &   0.363      &   0.006   &   0.509    &     0.962     &   0.606     &  0.011   &    0.586    &     0.981      &    0.686     &   0.009   &    0.637    &     0.963     &   0.642     &  0.005    &    0.604    &     0.693      &    0.135     &   0.006   &    0.726    &    0.765      &   0.339     &    0.018   \\
DL\_V3+ (s=8) \cite{chen2018encoder}        &    0.482    &     0.934      &    0.364     &   0.007   &    0.550   &     0.973     &    0.619    &  0.010   &   0.602     &     0.977      &    0.702     &  0.009    &   0.648     &     0.967     &   0.659     &   0.004   &    0.619    &    0.701       &    0.156     &   0.005   &    0.734    &    0.773   &   0.352     &   0.016  \\
DL\_V3+ (s=16) \cite{chen2018encoder}        &    0.502    &     0.920      &    0.375     &  0.005    &   0.558    &     0.972     &   0.624     &  \tb{0.008}   &   0.597     &    0.981       &    0.708     &   0.009   &    0.653    &     0.980     &   0.660     &  0.003   &    0.625    &     0.704      &    0.162     &   0.005  &    0.741    &     0.776     &   0.365   &  0.016  \\
U-Net  \cite{ronneberger2015u}        &    0.521    &      0.927     &    0.382     &   0.005   &    0.570   &    0.967      &   0.643     &   0.011  &    0.607    &     0.982      &      0.726   &   0.009   &    0.655    &     0.977     &   0.683    &   0.003   &   0.633     &     0.712      &      0.221   &    0.004  &    0.750    &     0.781     &   0.379     &   0.015  \\
Multi-class U-Net         &    0.515    &     0.923      &     0.389    &   0.005   &   0.547    &    0.967      &    0.647    &  0.010   &    0.618    &     0.982      &    0.731     &   0.010   &   0.649       &     0.976     &    0.685    &   0.004   &    0.631    &     0.709      &    0.223     &   0.004   &    0.748    &    0.779      &   0.383     &    0.015   \\
Attention U-Net  \cite{oktay2018attention}         &    \tb{0.536}    &     0.942      &     0.435    &   0.006   &    0.576   &     0.974     &    0.678    &  0.009   &  0.637      &     \tr{0.984}    &   0.762     &   \tb{0.007}   &    0.689    &    0.980      &    0.712    &    0.003   &    0.641    &     0.720      &     0.231    &   0.005   &   0.769     &    0.801   &   0.395   &  0.013    \\
Gated U-Net  \cite{schlemper2019attention}             &    0.529    &     \tb{0.945}      &    0.441     &  0.006    &   0.580    &    \tb{0.978}      &   0.682     &  0.009   &   0.638   &    \tb{0.983}    &     0.764    &   \tr{0.007}   &    0.685    &   0.982    &   0.716     &  \tb{0.003}   &    0.638    &     0.722      &   0.235    &   0.005   &   0.766     &    0.803    &  \tb{0.398}  &  \tb{0.013}      \\
Dense U-Net  \cite{li2018h}             &    \tr{0.559}    &     \tr{0.959}    &    \tr{0.469}     &   \tr{0.004}   &    \tr{0.617}   &     \tr{0.981}     &    \tr{0.697}    &   \tr{0.007}  &   \tr{0.649}     &  0.978         &    \tr{0.775}     &   0.008   &     \tr{0.723}   &    \tr{0.985}      &    \tr{0.726}    &   \tr{0.002}    &   \tr{0.649}     &     \tr{0.731}     &    \tr{0.245}    &   \tr{0.003}   &    \tr{0.781}    &     \tb{0.812}     &   \tr{0.403}     &     \tr{0.012}    \\
U-Net++  \cite{zhou2018unet++}          &    0.533    &    0.937      &    \tb{0.453}     &   \tb{0.005}   &    \tb{0.597}   &    0.974     &    \tb{0.689}    &   0.009  &    \tb{0.644}    &     0.980     &   \tb{0.771}     &   0.008   &    \tb{0.719}   &     \tb{0.984}     &    \tb{0.722}    &   0.003    &    \tb{0.645}    &    \tb{0.729}     &   \tb{0.241}    &  \tb{0.004}    &     \tb{0.777}   &    \tr{0.815}      &    0.397    &    0.013   \\ \hline
\end{tabular}
}
\end{table*}

\begin{figure*}[]
\begin{center}
\includegraphics[width=1.0\linewidth]{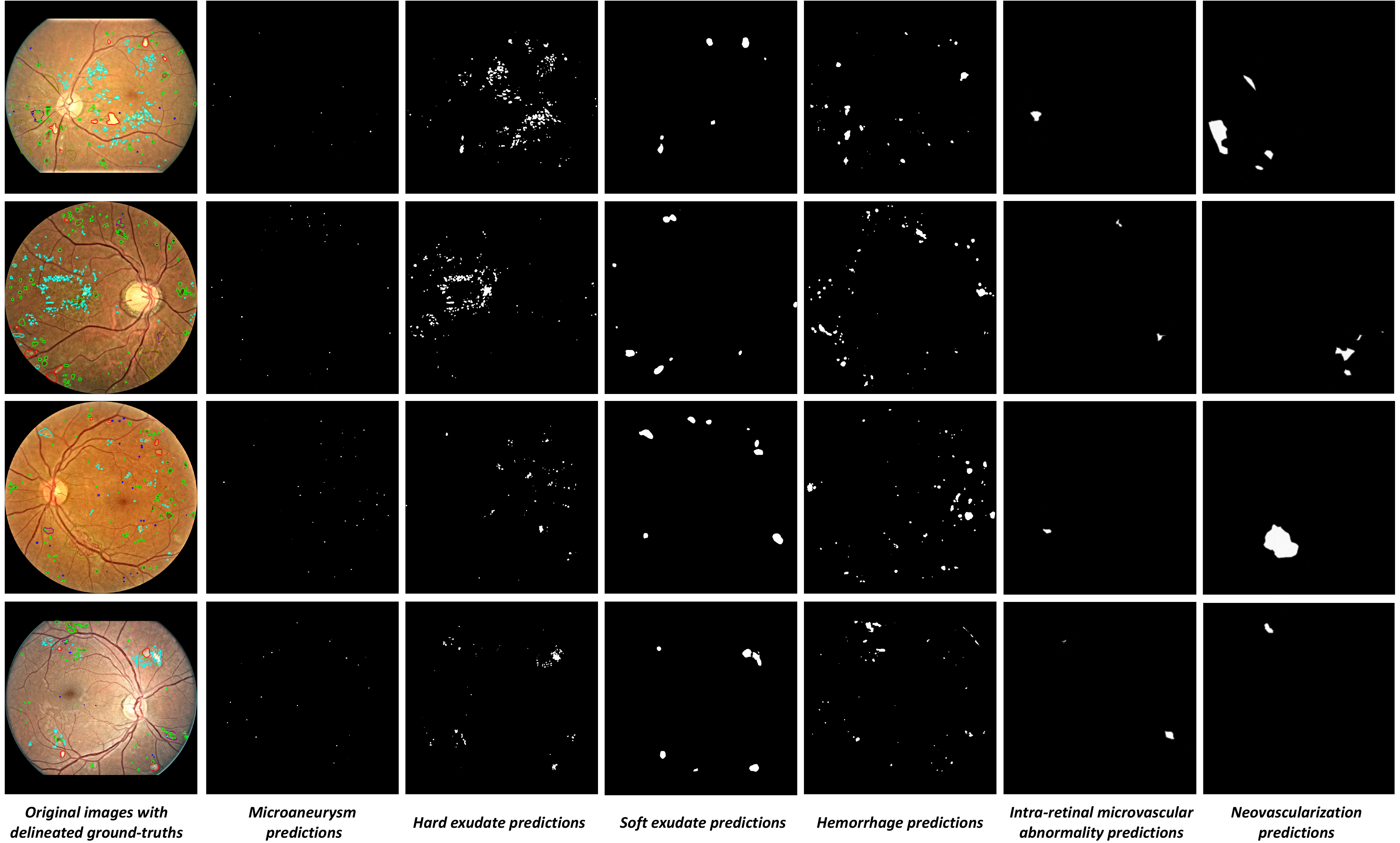}
\end{center}
   \caption{Qualitative segmentation performance of the Dense U-Net. All the mask outputs are binarized using a threshold of 0.25 for visualization.}
\label{fig:segmentation}
\end{figure*}

\section{Experiments and Results}

\subsection{Baselines}

\subsubsection{Segmentation}
To evaluate the DR lesion segmentation task, several classic semantic segmentation methods are adopted. They can be coarsely categorized into Non-U-Net frameworks and U-Net frameworks.

Non-U-Net Frameworks: FCN-8s \cite{long2015fully} employs a fully convolutional network which stacks multiple convolutional layers in an encoder-decoder fashion. The decoder upsamples the image using a transpose convolution to predict the segmented output. We use the setting of 8s to fuse the output. DeepLabV3+ \cite{chen2018encoder} also adopts the encoder-decoder architecture but introduces atrous spatial pyramid pooling, atrous separable convolution, and modified aligned Xception to enhance the performance. The settings of both $s=8$ and $s=16$ are tested.

U-Net Frameworks: U-Net \cite{ronneberger2015u} was proposed for biomedical image segmentation. Its most successful modification is to design a large number of feature channels with skip connections in the upsampling part, enabling the model to better propagate context information to higher resolutional layers. Multi-class U-Net is an extension that changes the binary output to multi-class outputs. Attention U-Net \cite{oktay2018attention} introduces end-to-end-trainable attention gates to separate localization and subsequent segmentation steps. This design can improve model sensitivity and accuracy to foreground pixels. Gated U-Net \cite{schlemper2019attention} was proposed with a novel attention gate to suppress irrelevant areas and focus on salient region features. Moreover, Dense U-Net \cite{li2018h} integrates a densely connected convolutional network into the U-Net framework, which strengthens the use of features and improves segmentation performance. U-Net++ \cite{zhou2018unet++} differs form the original U-Net in three ways - it has convolutional layers on skip pathways, has dense skip connections on skip pathways, and uses deep supervision, which enables model pruning. For all the baseline methods, single segmentation network is trained for each lesion, except Multi-class U-Net which six lesions share the backbone.

\subsubsection{Grading}
Task 2 is to rate the DR severity level from 0 to 4, which is a five-grade classification problem. We provide three kinds of baselines for evaluation. The \textbf{first} kind of baselines adopt a basic classification-only model with different classic backbones, including VGG-16 \cite{simonyan2014very}, ResNet-50 \cite{he2016deep}, Inception v3 \cite{szegedy2016rethinking}, and DenseNet-121 \cite{huang2017densely}. The \textbf{second} kind of baselines are ensemble models proposed by the top solutions in Kaggle competitions \cite{kaggle-eyepacs,kaggle-aptos2019}. The results of the various models are averaged to give a final prediction, which often yields substantial improvements in terms of accuracy. We adopt two baselines, denoted as Model Ensemble 1 and Model Ensemble 2 in Table~\ref{table:grading}. Model Ensemble 1 is the $1^{st}$ place solution of \cite{kaggle-eyepacs}, which combines three models - two convolutional networks using fractional max-pooling \cite{graham2014fractional} and a slightly modified VGG network. Model Ensemble 2 is the $1^{st}$ place solution of \cite{kaggle-aptos2019}, which consists of eight models, including Inception, ResNet, and SEResNeXt \cite{hu2018squeeze} variants. Last but not least, the \textbf{third} kind of baselines employ the idea of combining lesion identification and grading models. We assess three methods: the first one \cite{lin2018framework} learns lesion features using a visual attention model without pixel-level training, while the latter two \cite{Zhou_2019_CVPR,wu2020jcs} exploit lesion masks predicted from segmentation models to help grading. The backbones of \cite{Zhou_2019_CVPR,wu2020jcs} are both changed to DenseNet-121 for comparison.

 \subsubsection{Multi-label Classification}
 
 To evaluate the effectiveness of our proposed inductive transfer learning method for ocular multi-disease identification, we carry out two ablation studies. First, compared to a baseline which only adopts the basic classification module trained on the target domain data, the first ablation study is explores the effectiveness of the multi-scale transfer connections (Baseline+MTC). The multi-scale features learned from the source domain task are transferred to the target domain task. Moreover, the second ablation study validates that the adversarial domain-specific adaptation module (Baseline+MTC+DSAA) can improve the performance of the target domain task.
 
 The training scheme of our final Baseline+MTC+DSAA consists of two stages. In the first step, the segmentation module is pre-trained using the source domain data. The ADAM optimizer is adopted with a base learning rate of 0.01 and momentum of 0.5. We pre-train the segmentation module with a batch size of 32 for 100 epochs. In the second step, the two domain tasks are optimized together, along with the multi-scale transfer connections and domain-specific adversarial adaptation module. Hyper-parameters $\lambda$ and $\gamma$ are selected as 1 and 0.5, throughout experiments, which yields the best performance. The base learning rate is set to 0.001, and the batch size is set to 64. The training is completed after 300 epochs based on the target domain data length.

 \begin{table}[t]
\centering
\caption{\small Quantitative results of traditional lesion segmentation models. The best results are bolded.}
\label{table:seg_ma_he}
\resizebox{0.5\textwidth}{!}{
\begin{tabular}{c|ccccc}
\hline
Lesion              & Methods         & Dice  & ROC   & PR    & MAE   \\ \hline
\multirow{3}{*}{HE} & Splat+Pixel+KNN \cite{tang2012splat} & 0.504 & 0.957 & 0.581 & 0.013 \\
                    & Splat-wise+KNN \cite{tang2012splat} & 0.484 & 0.965 & 0.589 & 0.013 \\
                    & U-Net  \cite{ronneberger2015u}          & \textbf{0.570} & \textbf{0.967} & \textbf{0.643} & \textbf{0.011} \\ \hline
\multirow{3}{*}{MA} & MCF  \cite{zhang2009hierarchical}           & 0.486 & \textbf{0.942} & \textbf{0.385} & 0.006 \\
                    & FCN-8s  \cite{long2015fully}       & 0.468 & 0.925 & 0.363 & 0.006 \\
                    & U-Net  \cite{ronneberger2015u}          & \textbf{0.521} & 0.927 & 0.382 & \textbf{0.005} \\ \hline
\end{tabular}
}
\end{table}

\subsection{Results of Task: DR Lesion Segmentation}

In our experiments of lesion segmentation, the ratio of training and testing data is split as 1:1 for baseline comparisons. In each baseline method except Multi-class U-Net, different segmentation networks are trained for different lesion types. Table~\ref{tab:segmentation} provides the results of different methods, from which we can make the following observations. First, Dense U-Net and U-Net++ are the two best models for all lesions, except to segment hard exudates (EX) lesions, which no method obtained dominant performance on as these are relatively easy to segment. Second, the Multi-class U-Net shows a slight increase in AUC of PR compared to the standard U-Net, since all the lesions share the same model parameters to learn representations better. It significantly reduces the computational cost as well. Third, the U-Net frameworks obtain consistently better results than the non-U-Net frameworks, which demonstrates the advantages of the upsampling and skip connections of U-Net in allowing the network to propagate context information to higher resolution layers. Fourth, both of the attention modules proposed in Attention U-Net and Gated U-Net can significantly improve the segmentation performance compared to the basic U-Net. Last but not least, for microaneurysms (MA), intra-retinal microvascular abnormalities (IRMA), and neovascularization (NV), no current baseline model achieves satisfactory results. MAs are usually very tiny, and prone to being miss-detected or wrongly classified as hemorrhages (HE). The training data of IRMA and NV are still limited. Thus, better segmentation algorithms are expected to overcome these challenges in future research.

In addition to the deep segmentation frameworks, which can be adopted for all lesion detection tasks, some traditional classification methods have also been proposed to address one or two specific lesions related to DR. In \cite{tang2012splat}, a retinal hemorrhage detection method was introduced. It presents a method to extract splat features for splat-based hemorrhage detection. The feature extraction module includes splat features aggregated from pixel-based responses and splat-wise features. A filter and a wrapper approach are adopted in serious to select the features and reduce the dimensionality. K-nearest neighbor (KNN) searching is used to learn the classifier and obtain a hemorrhageness map. Moreover, to detect tiny-lesion MAs, traditional pixel classification methods can also work effectively since MAs can be encoded on low-level features. We evaluate \cite{zhang2009hierarchical}, which uses a multi-scale Bayesian correlation filter. In this approach, responses from a Gaussian filter bank are used to construct probability models of an object and its surroundings. When the responses of the correlation filtering are larger than a certain threshold, the detected locations are regarded as candidate microaneurysm locations. All the comparison results are shown in Table~\ref{table:seg_ma_he}.

\begin{table}[t]
\centering
\caption{\small DR grading results on EyePACS and FGADR. The best two results are shown in \tr{red} and \tb{blue} fonts.}
\label{table:grading}
\resizebox{0.5\textwidth}{!}{
\begin{tabular}{c|cc|cc}
\hline
Set                        &  \multicolumn{2}{c|}{EyePACS-test} & \multicolumn{2}{c}{FGADR-Grade-set} \\ \hline
Methods                       &    Acc. & Q.W.Kappa     &   Acc.     &  Q.W.Kappa   \\ \hline
VGG-16                        &     0.8363     &     0.8198     &     0.8043     &     0.7436   \\
ResNet-50                     &     0.8456     &     0.8239    &      0.8205    &   0.7576     \\
Inception v3                  &      0.8396    &      0.8111   &     0.8144     &    0.7493     \\
DenseNet-121                  &     0.8539     &     0.8349    &     0.8239     &    0.7678     \\
Model Ensemble 1                  &     0.8598     &    0.8482    &    0.8294      &    0.7737    \\
Model Ensemble 2                  &      0.8629    &     0.8521     &     0.8305     &    0.7786   \\  
Lin \cite{lin2018framework}      &    0.8671      &     0.8566   &      0.8362    &    0.7846     \\  \hline
Zhou \cite{Zhou_2019_CVPR} (DenseNet-121)  &      \tr{0.8945}     &      \tr{0.8846}      &     \tr{0.8603}     &    \tr{0.8482}     \\
Wu \cite{wu2020jcs} (DenseNet-121)                  &      \tb{0.8864}    &    \tb{0.8772}     &       \tb{0.8560}   &      \tb{0.8425}    \\ \hline
\end{tabular}
}
\end{table}

\begin{figure}[]
\begin{center}
\includegraphics[width=1.0\linewidth]{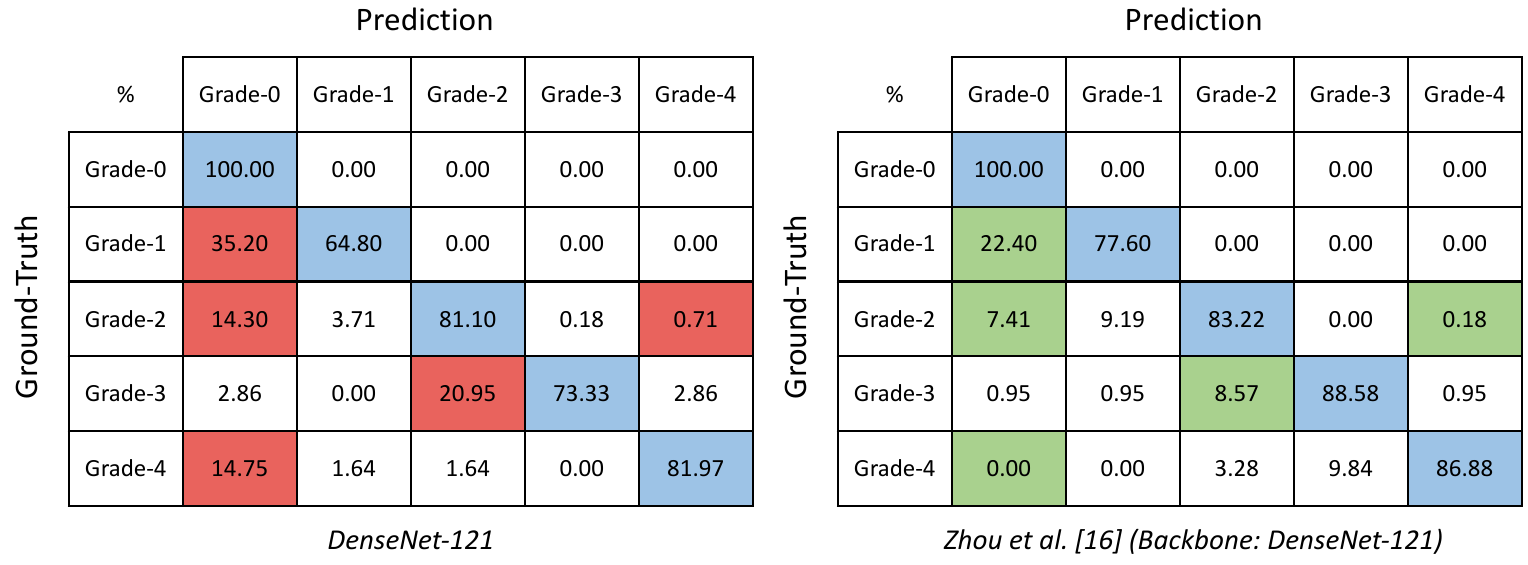}
\end{center}
   \caption{Comparison of the confusion matrices of DR grading (FGADR-Grade-set) between the methods with and without using lesion segmentation predictions. The blue blocks denote the correct predictions. From the red to green blocks, it clearly shows the decrease of incorrect grading results, by the help of segmented lesion masks.}
\label{fig:conf}
\end{figure}

\subsection{Results of Task 2: DR Grading}

We evaluate the grading results on both the test set of EyePACS (EyePACS-test) and the Grade-set of our FGADR (FGADR-Grade-set), as shown in Table~\ref{table:grading}, for a comprehensive comparison. First, the DenseNet-121 backbone achieves the best performance among the four individual models. Model ensemble further increases the result slightly. Moreover, although Lin \cite{lin2018framework} considered learning lesion attentions to help grading, the attention maps are learned in a weakly-supervised manner without pixel-level supervision. Thus, its improvement is limited. However, with the help of lesion masks predicted by fully-supervised segmentation models, notable improvements are obtained. Zhou \cite{Zhou_2019_CVPR} increases the Q.W.Kappa by 4.97\% and 8.04\% on the EyePACS-test set and FGADR-Grade-set, respectively. Wu \cite{wu2020jcs} increases the Q.W.Kappa by 4.23\% and 7.47\% on the two sets as well. For more details, we also provide a comparison of confusion matrices before and after using the lesion segmentation predictions in Figure~\ref{fig:conf}. As can be observed, the accuracies of classifying grade-1 and grade-3 are largely increased by 12.8\% and 15.25\%, respectively. The misclassification rate from grade-2 to grade-0 decreases by 6.89\%. Moreover, none of the grade-4 DR images are wrongly rated as Grade-0 or Grade-1 when the lesion masks are provided. Therefore, these improvements make DR diagnosis systems more robust and interpretable for ophthalmologists, since misclassification from high-severity DR levels to normal or early stage DR levels do not make sense.

\begin{table}[t]
\centering
\caption{\small Quantitative results of ocular multi-disease identification on ODIR-5K dataset.}
\label{table:classification}
\resizebox{0.42\textwidth}{!}{
\begin{tabular}{cccc}
\hline
Methods                    & Kappa & F-1 & ROC \\ \hline
VGG16                      &    0.5312   &  0.8892   &  0.8949   \\
Inception v3       &	0.6235   &   0.9054   &    0.9187    \\ 
Baseline (DenseNet)      &   0.6556    &    0.9163  &  0.9274    \\
Baseline+MTC            &   0.6843    &   0.9211  &   0.9316  \\
Baseline+MTC+AA &   \tb{0.7152}    &   \tb{0.9351}  &   \tb{0.9442}  \\ 
Baseline+MTC+DSAA &   \tr{0.7348}    &   \tr{0.9426}  &   \tr{0.9498}  \\ \hline
\end{tabular}
}
\end{table}

\begin{table}[t]
\centering
\caption{\small Accuracy of each ocular disease on ODIR-5K dataset. The best results are bolded.}
\label{table:acc_each_label}
\resizebox{0.50\textwidth}{!}{
\begin{tabular}{ccccc}
\hline
Ocular Diseases & B & B+MTC &  B+MTC+AA  & B+MTC+DSAA \\ \hline
Normal          & 0.8127   & 0.8465    &  0.8618   & \textbf{0.8699}              \\
\textbf{Diabetes}        & 0.8309   & 0.8505   &   0.8668   & \textbf{0.8735}              \\
Glaucoma        & 0.9776   & 0.9791  &   0.9831    & \textbf{0.9874}              \\
Cataract        & 0.9854   & 0.9863  &   0.9888    & \textbf{0.9906}              \\
\textbf{AMD}             & 0.9603   & 0.9731  &  0.9780     & \textbf{0.9826}              \\
\textbf{Hypertension}    & 0.9637   & 0.9751  &   0.9746    & \textbf{0.9788}              \\
Myopia          & 0.9923   & \textbf{0.9946} &   0.9938     & 0.9942              \\
Others          & 0.8538   & 0.8633  &   0.8770    & \textbf{0.8793}              \\ \hline
\end{tabular}
}
\end{table}

\begin{figure}[t]
\begin{center}
\includegraphics[width=1.0\linewidth]{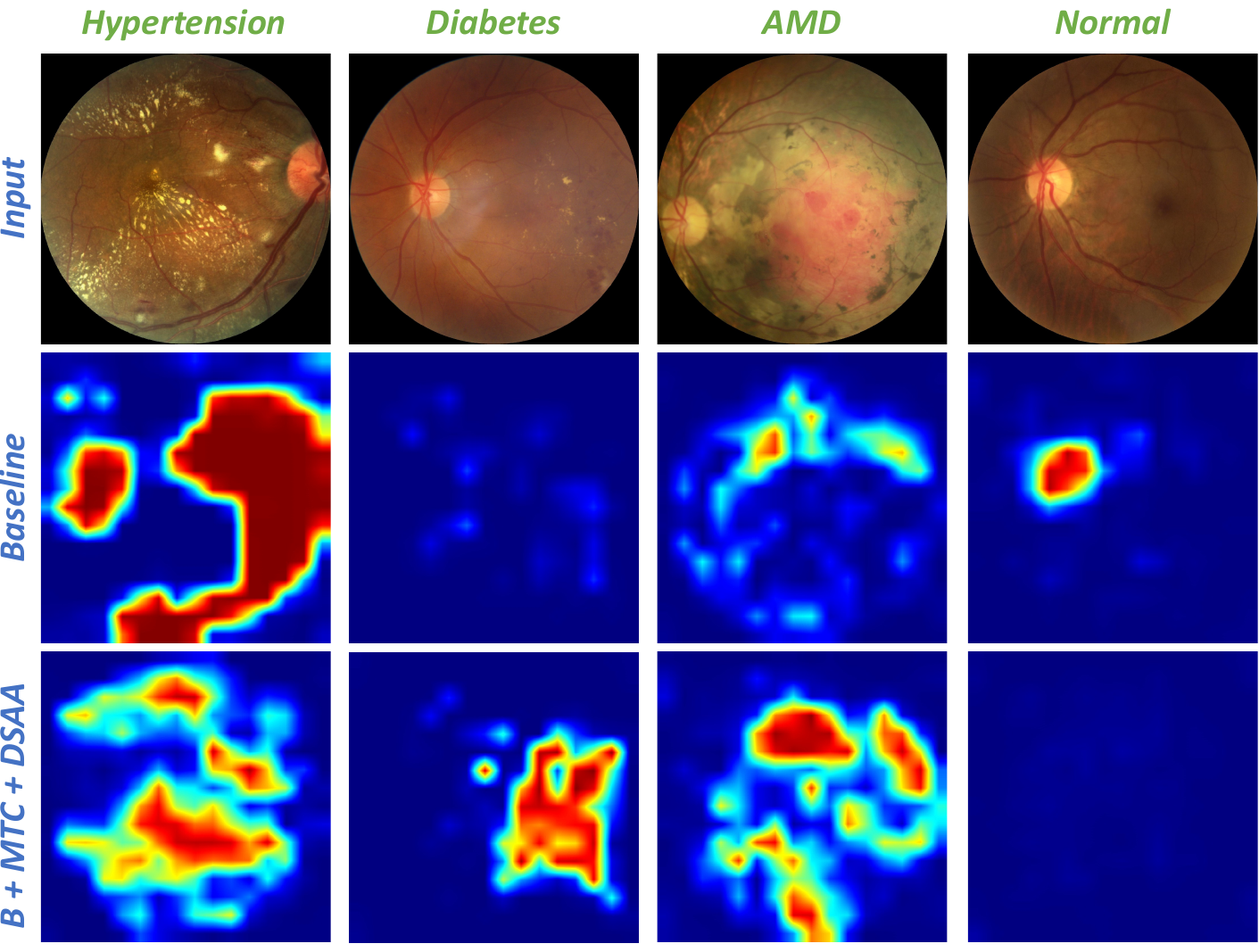}
\end{center}
   \caption{Visualization of the logit maps of the target domain network. The Baseline(B) and B+MTC+DSAA are compared.}
\label{fig:show}
\end{figure}

\subsection{Results of Task 3: Ocular Multi-Disease Identification}

To evaluate ocular multi-disease identification, 7000 images from the ODIR-5K \cite{odir2019} dataset are used for training and validation. Five-fold cross validation experiments are conducted. Table~\ref{table:classification} shows the results of different methods. We first evaluate individual models, VGG-16, Inception v3, and our DenseNet architecture, as baselines, where DenseNet achieves the best performance. Then, with the help of source domain task learning, the multi-scale transfer connections (MTC) increase the Kappa by 2.87\%. Moreover, the domain-specific adversarial adaptation (DSAA) module can further improve the model performance with an increase of 5.05\% in Kappa. The effectiveness of both designs have been validated. Compared to the normal adversarial adaptation (AA) which adopts the same $BN$ layers for the two domains, separate $BN$ layers of the domain-specific discriminator increase the Kappa by 1.96\%. For more details, the classification accuracies of each disease are illustrated in Table~\ref{table:acc_each_label}. We observe that the transfer learning from our fine-grained annotated DR domain data can consistently improve the identification results for all the ocular diseases in the task domain. Particularly, for diabetes, AMD, and hypertension, the improvements are significant, while slight gains are achieved for glaucoma, cataracts, and myopia. To better interpret the effectiveness of transfer learning from the source domain to target domain, we visualize the final logit maps of the samples correctly classified by our transfer learning method but wrongly classified by the baseline model. As illustrated in Figure~\ref{fig:show}, we observe that the logit maps extracted by Baseline+MTC+DSAA can contain more precise lesion regions related to the disease, because the lesion segmentation ability learned from the source domain network is integrated into the target domain network.

\section{Conclusion}
To promote research in medical image segmentation, classification, and transfer learning, particularly for the community of diabetic retinopathy diagnosis, in this paper, we proposed a large fine-grained annotated DR dataset, FGADR. Moreover, we conducted extensive experiments to compare different state-of-the-art segmentation models and explore the lesion segmentation task. Joint classification and segmentation methods were demonstrated to have better performance on the DR grading task. We also developed an inductive transfer learning method, DSAA, to exploit our DR dataset for improving ocular multi-disease identification.

{
\bibliographystyle{IEEEtran}
\bibliography{DR_Benchmark.bib}
}

%
\IEEEpeerreviewmaketitle

\end{document}